\documentclass[twocolumn, switch]{article} 

\usepackage{preprint}
\usepackage[ruled,vlined]{algorithm2e}
\usepackage{algorithmic}
\usepackage{amsmath, amsthm, amssymb, amsfonts}
\usepackage{tabularx}
\usepackage{adjustbox}
\usepackage{multirow}

\usepackage{graphicx}

\usepackage[numbers,square]{natbib}
\usepackage[utf8]{inputenc}	
\usepackage[T1]{fontenc}	
\usepackage{xcolor}		
\usepackage[colorlinks = true,
linkcolor = purple,
urlcolor  = blue,
citecolor = cyan,
anchorcolor = black]{hyperref}	
\usepackage{booktabs} 		
\usepackage{nicefrac}		
\usepackage{microtype}		
\usepackage{lineno}		
\usepackage{float}			

\usepackage{lipsum}		

\usepackage{newfloat}
\DeclareFloatingEnvironment[name={Supplementary Figure}]{suppfigure}
\usepackage{sidecap}
\sidecaptionvpos{figure}{c}

\usepackage{titlesec}
\titlespacing\section{0pt}{12pt plus 3pt minus 3pt}{1pt plus 1pt minus 1pt}
\titlespacing\subsection{0pt}{10pt plus 3pt minus 3pt}{1pt plus 1pt minus 1pt}
\titlespacing\subsubsection{0pt}{8pt plus 3pt minus 3pt}{1pt plus 1pt minus 1pt}

\title{FC-MIR: A Mobile Screen Awareness Framework for Intent-Aware Recommendation based on Frame-Compressed Multimodal Trajectory Reasoning}

\usepackage{eso-pic}
\usepackage{tikz}
\usepackage{xcolor}
\PassOptionsToPackage{colorlinks=true,linkcolor=gray,urlcolor=gray}{hyperref}

\usepackage{titling}
\usepackage{orcidlink}
\usepackage{footmisc}
\newcommand{\Author}[2]{
	\textbf{#1}\textsuperscript{#2}%
}

\author{
	\Author{Zhe Yang}{1,2},
	\Author{Xiaoshuang Sheng}{1},
	\Author{Zhengnan Zhang}{1},
	\Author{Jidong Wu}{1},
	\Author{Zexing Wang}{1},\\
	\Author{Xin He}{1},
	\Author{Shenghua Xu}{1},
	\Author{Guanjing Xiong}{1}
}

\date{%
	\textsuperscript{1}vivo AI Lab \\
	\textsuperscript{2}Zhejiang University\\[1em]
	\footnotesize \textbf{Corresponding author:} Xiaoshuang Sheng \texttt{shengxiaoshuang@vivo.com}\\
}

\begin{document}
	
	\twocolumn[ 
	
	\maketitle
	\thispagestyle{empty}

	\vspace{0.35cm}
	
	] 
	

	\section*{abstract}
	Identifying user intent from User Interface (UI) operation trajectories stands as a pivotal challenge in achieving comprehensive UI understanding and enabling mobile task automation agents. While Multimodal Large Language Models (MLLMs) have recently demonstrated remarkable proficiency in tasks such as video understanding, their application in real-time mobile inference and lightweight deployment is hindered by high computational overhead and the lack of efficient redundant frame processing mechanisms.
	To mitigate these limitations, we propose the FC-MIR framework. By utilizing keyframe sampling and adaptive concatenation strategies, FC-MIR effectively reduces visual redundancy and enhances inference efficiency. It integrates state-of-the-art closed-source MLLMs or fine-tuned MLLMs (Qwen3-VL) to perform user trajectory summarization and intent prediction. Building upon this, we further extend the task boundaries to investigate the feasibility of generating post-prediction operations and search suggestions. We also introduce a novel fine-grained evaluation metric designed to assess the practical utility of the generated summaries, predictions, and suggestions.
	To facilitate rigorous evaluation, we construct a UI trajectory dataset comprising scenarios automatically generated by UI-Agents (Agent-I) and real-world user interactions (Person-I). Experimental results demonstrate that our compression methodology preserves performance integrity even at compression rates of 50\%-60\%. Both closed-source and fine-tuned MLLMs exhibit robust intent summarization capabilities, indicating the potential for lightweight on-device deployment. However, regarding suggestions, MLLMs still face challenges related to usefulness and the "surprise" factor, highlighting significant room for improvement in this domain. Finally, we deploy our framework in a real-world setting, achieving the integration of UI perception and UI-Agent proxies, thereby laying a solid foundation for further exploration and development in this field.

	\section{Introduction}
	With the increasing use of smart devices (especially mobile devices) in daily tasks,the significance of interaction has become increasingly prominent. The user interface (UI) remains the primary medium through which humans interact with applications, whether directly or through agents. Consequently, accurately perceiving UI operation trajectories and summarizing user intent can significantly optimize the agency process, thereby facilitating the realization of mobile task automation agents \cite{fu2025ui,nguyen2025gui}.
	
	Consider the scenario of a user booking flight tickets for a vacation: an ideal agent would observe and analyze these interactions, understand the user's underlying intent, and then proactively offer suggestions such as booking hotels for the same period, scheduling these dates in the calendar, or providing travel itineraries for the destination.
	
	However, understanding UI trajectories faces significant challenges. For instance, the visual presentation of different applications varies widely; UI interfaces are not static—they dynamically change in response to user actions, environmental factors, or backend logic, which directly disrupts the completeness and coherence of the trajectories. In practical interactions, user trajectories inevitably contain a large amount of redundant or ineffective information. Additionally, the same UI trajectory may correspond to multiple potential intents, and these intents can exist at different levels of abstraction.
	
	An efficient perception model should be capable of understanding users' interaction trajectories and underlying goals or intents, allowing the agent to proactively adjust its behavior to match user preferences, habits, and long-term objectives. This capability is a crucial step toward personalization, enabling the agent to provide context-aware, efficient assistance tailored to individual user needs \cite{li2024personal}.
	
	In recent years, Multimodal Large Language Models (MLLMs) have achieved significant strides in comprehending complex video content \cite{qwen3technicalreport,zhang2025videollama}, delivering state-of-the-art performance across various tasks, including video understanding, video question answering, and detailed video captioning \cite{Qwen2.5-VL,wang2025vdcagentvideodetailedcaptioners,comanici2025gemini}. Nevertheless, processing videos remains computationally expensive: a large number of video frames need to be input into the language model along with text prompts, and each frame, after being encoded by a visual encoder, generates thousands of tokens. This not only leads to substantial computational overhead but also introduces significant visual redundancy. 
	Furthermore, the processing of redundant video frames is ill-suited for rapid on-device inference and the swift execution of downstream UI automation agents  \cite{xiao2025ui,wang2024mobile,nguyen2025gui}. Therefore,there is an urgent need for a lightweight solution that effectively reduces redundant visual tokens while maintaining comparable accuracy in mobile user intent recognition tasks. 
	Current research has been devoted to scaling down multimodal models \cite{fu2025ui}, but their sizes (\textasciitilde 4.4B parameters) are still far from the optimal level required for stable operation on high-end mobile devices. Additionally, training these models consumes vast amounts of data and computational resources.
	
	Some existing works have explored tasks such as user intent recognition on PC and mobile platforms.  
	Studies like ScreenLLM and UI-tra \cite{jin2025screenllm,berkovitch2025identifying} have investigated the application of Vision-Language Models (VLMs) in interface understanding, utilizing pre-trained models to capture the correlations between interface elements and user interactions.  
	Works such as UI-JEPA \cite{fu2025ui} aim to construct self-supervised frameworks for interface representation learning, offering novel technical pathways for user behavior analysis.
	
	However, these approaches exhibit limitations regarding multimodal data fusion and real-time analysis within real-world application scenarios.   
	ScreenLLM primarily targets specific software rather than real-life usage contexts.  
	UI-JEPA still underperforms compared to mainstream large models across various metrics. Although it can handle some few-shot tasks, it struggles to achieve satisfactory results in zero-shot settings, making it unsuitable for complex scenarios. Moreover, it requires substantial data and computational resources for training, hindering practical deployment on mobile devices.  
	UI-tra provides a paradigm for UI trajectory intent prediction, but the trajectories are predominantly single-task completion types. They do not encompass real-world complexities such as multi-task interleaving (e.g., searching for hotels while booking flights), dynamic goal changes, or operational errors. Additionally, UI-tra does not process sequential frames directly, relying instead on pre-processed standard datasets. These datasets are ill-suited for complex real-world scenarios characterized by redundancy, diversity, and multi-intent interactions, leaving significant room for improvement in intent prediction accuracy.
	
	To address the aforementioned challenges and limitations, and to ensure the accuracy of intent analysis, we optimize on-device algorithms for UI trajectory extraction. This approach resolves issues related to the speed and accuracy of video intent prediction. Furthermore, we extend the task boundaries to explore the feasibility of search and operation recommendations based on intent recognition analysis. In this paper, we propose an MLLM-based method for understanding UI video trajectory behaviors. By employing innovative sequential keyframe selection and concatenation strategies, our method achieves deep comprehension of user interface trajectories and accurate intent recognition, subsequently providing search and operation suggestions to enable Agent automation.
	Our main contributions are summarized as follows:
	
	1. A Novel Agent Framework: We validate the feasibility of using MLLMs for UI trajectory analysis, intent recognition, and agent automation, establishing an end-to-end pipeline from screen state capture to automated task completion. This diversifies agent inputs beyond natural language.
	
	2. Train-Free Trajectory Extraction: We propose an adaptive, training-free keyframe sampling and stitching algorithm for video sequences, which significantly reduces frame redundancy while preserving critical trajectory information, thereby enhancing both the speed and accuracy of model inference.
	
	3. Lightweight Deployment Exploration: Using reinforcement learning (GRPO), we further explore the capabilities and potential improvements of lightweight models in intent recognition and prediction tasks, providing valuable references for future on-device deployment.
	
	4. Benchmark Datasets: We release Chinese UI trajectory intent recognition datasets (Agent-I and Person-I) tailored for mobile environments. These datasets expand the linguistic context, cover real-world usage scenarios, and have been rigorously evaluated across multiple MLLMs through both human and LLM-based assessments.
	
	\section{Background and Related Work}
	\label{sec:headings}
	
	\subsection{UI-understanding}
	Several machine learning models have been proposed to enhance UI understanding. Early studies \cite{bai2021uibert,he2021actionbert,baechler2024screenai} primarily focused on:  
	leveraging large models to generate large-scale unlabeled UI data for pre-training Transformer-based models to learn general feature representations at the UI component level.  
	Other methods \cite{zhang2021screen} incorporated semantic information and heuristic rules during model training to improve UI detection performance. However, these approaches often have limitations: they target only individual UI components, making it difficult to comprehend "task concepts" or learn comprehensive visual representations. More recent methods \cite{jin2025screenllm} enhance the extraction and understanding of UI elements and cursors through pixel change detection and lightweight CNN models, but such approaches are tailored to specific software, exhibit poor scalability in large-scale multi-task scenarios, and have limited generalization to unseen tasks.  
	The method in \cite{zhang2025summact} addresses the issue of LLMs overlooking critical UI details by incorporating attention weights into the loss function, forcing the model to focus on UI element content. However, it relies on textual descriptions of UI elements (e.g., HTML structure) and does not leverage visual cues from GUI screenshots (such as icons and layout), resulting in limited capability for handling text-free UIs (e.g., icon-only buttons).
	
	In contrast to the above studies, our approach aligns with methods like \cite{berkovitch2025identifying,fu2025ui} by treating video as the processing object—these videos record sequences of UI operations during task execution. There are three main reasons for choosing video as the primary input:  
	1. Video is the predominant medium for monitoring user activities on mobile devices.  
	2. Although additional inputs like structured metadata could be beneficial, such information is not always accessible from third-party applications in real-world scenarios.  
	3. Videos contain richer information than static images and can preserve causal relationships.
	
	Furthermore, with the rapid evolution of applications and UI layouts, models trained on specific UI elements and hierarchies struggle to generalize effectively across different application versions. Our method addresses this by optimizing keyframe selection through a specific algorithm and then combining it with a post-trained MLLM to convert the input into a textual representation of user intent. This approach not only retains the temporal dynamics of UI interactions but also enables a more comprehensive understanding of unseen user tasks.
	
	\subsubsection{Large Vision-Language Models for Video}
	Transformer-based large language models (LLMs) have revolutionized natural language processing \cite{dubey2024llama,Qwen2VL}. Researchers have extended the multimodal processing capabilities of LLMs by incorporating multimodal inputs, typically visual content such as images and videos \cite{li2024llava,vasu2025fastvlm}, to build large vision-language models (VLMs). These models leverage large-scale pre-training to learn rich representations that capture the complex relationships between visual and textual modalities.
	
	Building on this foundation, large VLMs have significantly advanced various video understanding tasks, including detailed video description, video captioning \cite{chai2024auroracap,wang2025vdcagentvideodetailedcaptioners,chen2024sharegpt4video}, video question answering \cite{yang2025svbench,fei2024video,min2024morevqa}, and spatio-temporal reasoning \cite{wang2025lastlearningthinkspace,qian2024momentor,chen2024rextime}. Recent surveys \cite{li2025benchmark,liang2024survey,tang2025video} provide comprehensive reviews of VLM architectures, training strategies, and applications in video understanding, highlighting both their capabilities and the challenges they face.
	
	Despite these successes, large VLMs still face significant challenges when processing videos. These challenges stem from the large number of video frames, the limited context length of language models \cite{wang2025mmlongbench,wu2024longvideobench}, and insufficient retrieval and localization of key information \cite{hu2025nemo}. These limitations necessitate efficient frame selection strategies to identify and retain essential information within videos without overwhelming the model. Therefore, developing effective frame selection methods is crucial for further enhancing the performance of VLMs in video understanding tasks.
	
	\subsection{Video Token Reduction in Large VLMs}
	Reducing video tokens is crucial for the computational feasibility of large Vision-Language Models (VLMs). The most common and straightforward method is uniform frame sampling, which selects frames without considering their importance to the task. However, this approach may miss key frames necessary for understanding complex or long videos, thereby limiting reasoning performance. Several strategies have been proposed in the literature to make token reduction more adaptive and task-specific.
	
	\paragraph{Token Merging}  
	One approach involves merging information within or across frames to form compact yet comprehensive representations. GRT \cite{zhang2025dense} employs a two-stage framework that uses pixel-level gating and semantic-level scene merging to accelerate and reduce tokens. Koala \cite{tan2024koala} aggregates local segment information and inter-segment relationships through conditioned segments and a video tokenizer, while LongVU \cite{shen2024longvu} leverages cross-modal queries and inter-frame dependencies to reduce redundancy while preserving visual details. However, token merging methods may oversimplify fine-grained information, potentially causing the model to miss subtle differences between frames.
	
	\paragraph{Query-Based Frame Sampling} 
	To overcome the limitations of uniform sampling, recent similarity-based methods select frames based on their relevance to a specific query or task \cite{park2024too,tang2025adaptive,liu2025bolt}. For example, BOLT \cite{liu2025bolt} uses inverse transform sampling (ITS) as a frame selection method, relying on cosine similarity between CLIP\cite{radford2021learning} query embeddings and captions to sample key frames. VQOS enhances similarity matching by adding a candidate selection module. Although these methods are training-free, the computation time for similarity measures can be comparable to inference time. Other approaches learn optimal frame selection strategies through additional training. For instance, Frame-Voyager \cite{yu2024frame} learns informative frame combinations by training on datasets where frames are ranked based on VLM prediction loss. While such learnable sampling methods may be more effective, they require additional training and substantial computational resources, limiting their practicality.
	
	In contrast to the above methods, this paper proposes a training-free, pixel-level frame selection strategy along with an adaptive stitching algorithm. This approach aims to minimize redundancy in videos while preserving trajectory information, reducing the burden on VLMs during inference and enhancing their performance in understanding tasks.

	\section{Sequence frame sequence intent prediction framework}
	\subsection{Overview}
	
	The overall architecture of our method is illustrated in the Figure 1. First, we sample a large number of frames from the video to ensure sufficient visual coverage.  
	Next, we apply a pixel-level compression strategy to filter out blurry and redundant frames, retaining those with richer semantic information.  
	To further eliminate redundancy, we employ an adaptive stitching compression strategy, resulting in a more compact yet informative image.  
	Subsequently, the selected frames are fed into a large Multimodal Large Language Model (MLLM) to perform tasks such as trajectory understanding, intent summarization and prediction, and recommendation.  
	The intent prediction task will provide suggestions for the user to choose from. The user only needs to select the task of interest to initiate downstream activities such as UI automation agency or in-depth search.
	
	\begin{figure*}[t] 
		\centering 
		\includegraphics[width=\textwidth, keepaspectratio]{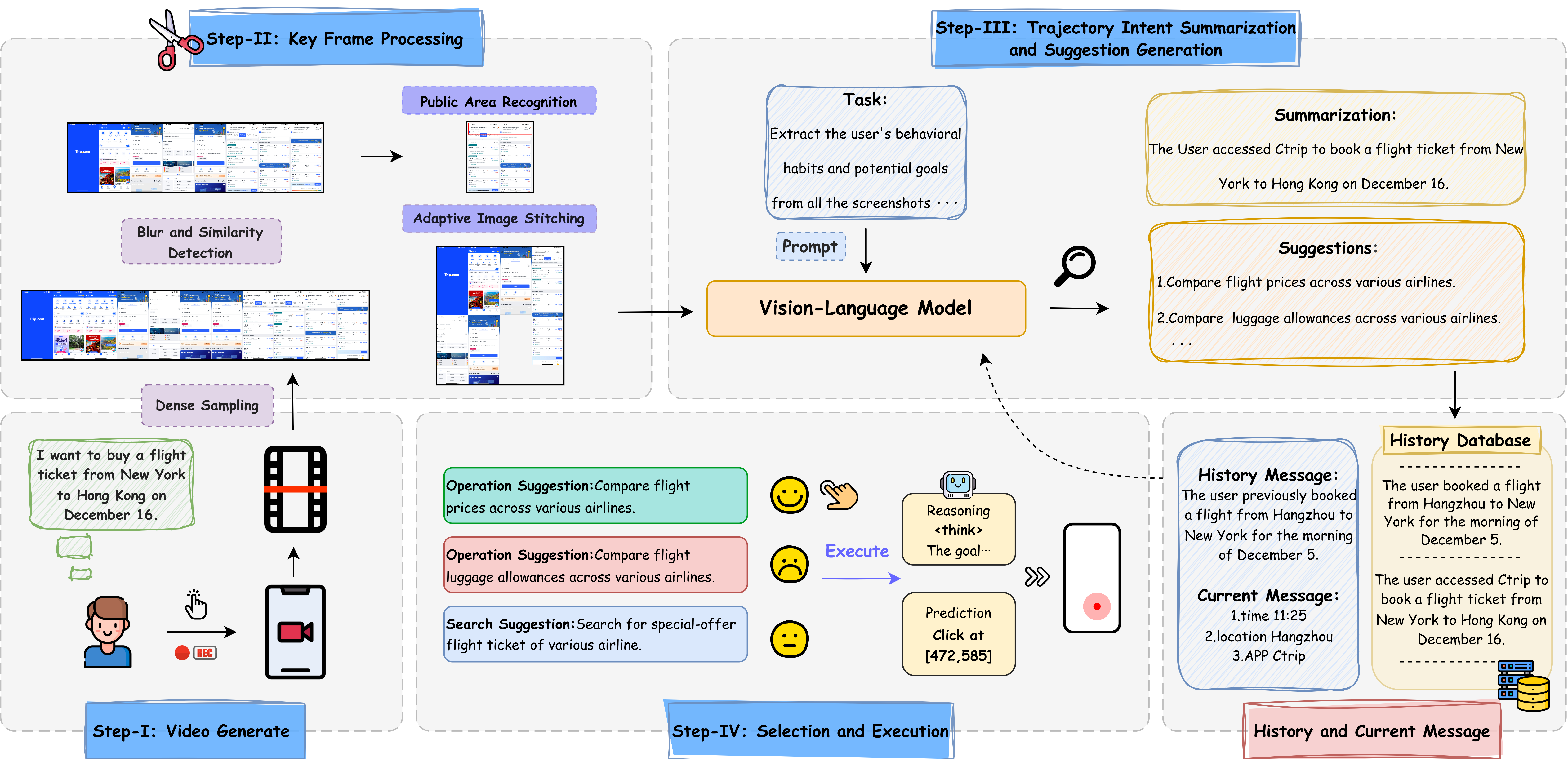}
		\caption{Overview:
			(I) Video Generation:Initiated by the user to start capturing on-device status.  
			(II) Key Frame Processing:Achieves video frame compression through blur detection, similarity detection, and adaptive stitching.  
			(III) Trajectory Intent Summarization and Suggestion Generation:Transmits key sequence frames and text information into an MLLM to generate user intent predictions and suggestions.  
			(IV) Selection and Execution:The user selects the desired suggestions, which are then passed to a downstream UI-Agent for execution.}
		\label{fig:flowchart}
	\end{figure*}
	
	\subsection{Algorithms}
	To address the specific requirements of processing screenshot sequences, we propose a multi-stage framework for keyframe selection and adaptive stitching. This framework achieves efficient keyframe perception and seamless image stitching, providing reliable technical support for user intent perception and automated testing.
	
	\subsubsection{Keyframe Selection}
	Keyframe selection is the core step in sequential frame processing. It filters out redundant or low-quality image content, retaining only the essential information from key frames.
	
	\paragraph{Blur Detection}
	Given that videos are prone to motion blur, our blur detection method is based on variance analysis using the Laplacian operator. It assesses image clarity by calculating the variance of the Laplacian of the grayscale version of the image.
	The Laplacian operator is a second-order differential operator capable of detecting edges and details in an image. For a grayscale image \(I(x,y)\), the Laplacian is defined as:
	
	\[
	\nabla^2 I = \frac{\partial^2 I}{\partial x^2} + \frac{\partial^2 I}{\partial y^2}
	\]
	The variance is calculated as:
	
	\[
	\sigma^2 = \frac{1}{N}\sum_{i=1}^{N}(L_i - \mu)^2
	\]
	where \(L_i\) represents the Laplacian response value, \(\mu\) is the mean, and \(N\) is the total number of pixels.
	Through experimental validation, we set the blur detection threshold \(\Gamma\) to 100. This value, determined based on statistical analysis of a large set of test images, effectively filters out blurry images while preserving clear ones.
	
	\paragraph{Similarity Detection}
	We employ a two-stage similarity detection approach, combining global and local analysis. For the global stage, we use the Perceptual Hash algorithm \cite{farid2021overview}, which rapidly performs grayscale conversion, DCT transformation, frequency domain filtering, and quantization. Image similarity is checked by comparing the Hamming distance of the hash values.
	
	Given the dense information typically present on mobile screens, outright deletion of certain views might lead to task failure. Therefore, we utilize a sliding window Structural Similarity Index Measure (SSIM) \cite{wang2004image} for local, partially overlapping detection. SSIM measures image similarity based on luminance, contrast, and structure:
	
	\[
	\text{SSIM}(x,y) = \frac{(2\mu_x\mu_y + C_1)(2\sigma_{xy} + C_2)}{(\mu_x^2 + \mu_y^2 + C_1)(\sigma_x^2 + \sigma_y^2 + C_2)}
	\]
	where \(\mu_x, \mu_y\) are local means, \(\sigma_x^2, \sigma_y^2\) are local variances, \(\sigma_{xy}\) is the local covariance, and \(C_1, C_2\) are constants for stability.
	To accelerate deployment on mobile devices, we introduce optimization strategies such as downsampling, histogram pre-screening, and adaptive windows.

	\subsubsection{Adaptive Image Stitching}
	We observed that mobile videos, compared to PC videos, contain a significant amount of scrolling operations. Retaining all frames globally would lead to information redundancy. Therefore, we employ a stitching algorithm to ensure both information completeness and continuity.
	We utilize a perceptual hash-based row-block similarity detection to identify and remove common areas in the images (such as the status bar and navigation bar), which are then integrated only after the final stitching is complete.
	
	We employ the ORB algorithm \cite{rublee2011orb}, which combines FAST corner detection and BRIEF descriptors, offering rotation invariance and high computational efficiency.
	For two descriptors \(d_1\) and \(d_2\), we measure their dissimilarity using the Hamming distance:
	\[
	d_H(d_1, d_2) = \sum_{i=0}^{n-1} |b_{1,i} - b_{2,i}|
	\]
	Subsequently, we use K-Nearest Neighbors (KNN) \cite{cover1967nearest} to find the \(k\) nearest neighbor matches for each descriptor. In our experiments, we set \(k=2\), meaning we find the best and the second-best matches.
	
	Based on Lowe's mathematical assumption \cite{lowe2004distinctive}: the distance distribution of correct matches should cluster near smaller values, while the distribution for incorrect matches is more scattered.
	For the best match \(m_1\) and the second-best match \(m_2\), if
	\[
	\text{ratio} = \frac{d_H(m_1)}{d_H(m_2)} < \tau
	\]
	then the match is considered correct, where \(\tau = 0.5\) is the ratio threshold.
	
	For overlap position calculation, we use the median method. Given a set of matching points \(\{(x_i, y_i, x_i', y_i')\}_{i=1}^N\), the overlap position is calculated as:
	\[
	y_{\text{pos}} = \operatorname{median}_{i=1}^N \left( y_i - y_i' \right)
	\]
	This effectively ensures the stability of the stitching.

	\begin{algorithm}[]
		\caption{Similarity-based Video Frame Sampling}
		\KwIn{Video $V$, interval $\Delta t$, 
			$T_{ph}$, $T_{ssim}$,  $B_{thr}$}
		\KwOut{Retained frames $\{F_k\}$}

		skip $\leftarrow \lfloor$fps $\cdot \Delta t\rfloor$\;
		prev $\leftarrow$ None; batch $\leftarrow [\,]$\;

		$i \leftarrow$ current frame index\;
		\If{$i \bmod$ skip $= 0$}{
			
			\If{ not Blur($I$, $B_{thr}$)}{
				
				\If{prev is None}{
					batch $\leftarrow [I]$\;
				}
				\Else{
					sim $\leftarrow$ HybridSimilar(prev, $I$, $T_{ph}$, $T_{ssim}$)\;
					\If{sim = False}{
						\If{batch not empty}{
							append last(batch) to $F$
						}
						batch $\leftarrow [I]$\;
					}
					\Else{
						append $I$ to batch\;
					}
				}
			}
			prev $\leftarrow I$\;
		}
		
		\If{batch not empty}{
			append last(batch) to $F$
		}
		
		\Return $F$
	\end{algorithm}

	\begin{algorithm}[]
		\caption{Batch Screenshot Stitching}
		\KwIn{Ordered frames $\mathcal{F} = \{f_0, f_1, \ldots, f_{n-1}\}$}
		\KwOut{Stitched images $\mathcal{S}$}
		
		$\mathcal{S} \leftarrow \emptyset$, $I_{acc} \leftarrow$ None, $i \leftarrow 0$\;
		
		\While{$i < |\mathcal{F}|$}{
			\eIf{$I_{acc} = $ None}{
				$I_{acc} \leftarrow f_i$, $i \leftarrow i+1$\;
			}{
				\tcp{Detect and remove common UI bars}
				$h_{top}, h_{bot} \leftarrow$ DetectCommonBars($I_{acc}$, $f_i$)\;
				
				$I'_{acc},f'_i \leftarrow$ RemoveBars($I_{acc}$, $f_i$)\;

				\tcp{Find overlap using ORB features}
				$I_{stitch} \leftarrow$ ORBStitch($I'_{acc}$, $f'_i$)\;
				
				\eIf{$I_{stitch} \neq $ None}{
					$I_{acc} \leftarrow$ AddBars($I_{stitch}$, $h_{top}$, $h_{bot}$)\;
					$i \leftarrow i+1$\;
				}{
					
					$\mathcal{S}$.append($I_{acc}$)\;
					$I_{acc} \leftarrow$ None\;
				}
			}
		}
		
		\If{$I_{acc} \neq $ None}{
			$\mathcal{S}$.append($I_{acc}$)\;
		}
		
		\Return $\mathcal{S}$
	\end{algorithm}
	
	\section{Definition and Evaluation Methodology}
	This section outlines our task definition and evaluation methodology. Given an input UI trajectory and its corresponding gold task description, we will evaluate the predicted task description, various metrics for intent recognition, and the utility of subsequent intent prediction and recommendations. This section first provides necessary definitions and then introduces the evaluation metrics.
	
	\subsection{Task Definition}
	Previous work has primarily focused on inferring the user's original intent from an observable UI trajectory generated while the user attempts to complete a specific task. As mentioned in \cite{berkovitch2025identifying}, this scenario is essentially the inverse problem of a known "UI automation task." These studies often repurpose UI automation datasets by reversing the roles of the components for testing. However, this approach suffers from idealized assumptions (e.g., no frame redundancy, complete trajectories, bounding box annotations) and limited scenarios, primarily within English-language environments.
	
	Our task differs from these deterministic UI trajectory tasks. For instance, \cite{berkovitch2025identifying} predicts outcomes based on a complete and concise chain of trajectories, which are relatively simple, complete, and involve single-app interactions. Our scenario focuses on summarizing the trajectory and intent from a series of user UI actions, and providing predictions for the user's next operation or search suggestion, even when the trajectory is incomplete.
	
	Consider an example: suppose the user proactively enables screen state awareness and captures a short video while on a chat app interface. In the video, the on-screen chat history mentions two different restaurants and includes a snippet of conversation. This constitutes an incomplete UI trajectory, because the user might subsequently proceed to view information such as the restaurants’ locations or prices. Our goal is to predict the user's next action based on the UI trajectory, offering timely operation or search suggestions to facilitate subsequent agent-based automation. We observe a strong correlation between the quality of the UI trajectory summary (i.e., summarizing the prior intent) and the accuracy of predicting the next intent. Therefore, we incorporate a summarization task, which can also aid in building a unique user memory.

	\subsection{Evaluation Metrics}
	
	Previous work primarily focused on the satisfaction relationship between predicted descriptions and ground truth descriptions \cite{berkovitch2025identifying,caduri2025bi} or used hard metrics (SBERT, ROUGE) for measurement \cite{fu2025ui}. These approaches essentially provide a coarse-grained evaluation of user trajectory summarization and do not assess the ability to understand the core intent and plan subsequent tasks. Therefore, we introduce more fine-grained evaluation criteria to validate the feasibility of this task.
	
	For the prior intent summarization, we employ five metrics for evaluation:
	
	\begin{itemize}
		\item Action Information Completeness: Whether the model's summary includes the completeness of each operation, containing all necessary actions.
		\item Action Sequence Accuracy: Whether the sequence of operations in the model's summary matches the reference summary.
		\item Object Detail Accuracy: Whether the model's summary explicitly states specific details of the operation objects (e.g., App names, filter conditions, search keywords, news headlines) instead of vague generalizations, and does not contain incorrect details.
		\item Output Format Standardization: Whether the model's summary is a single-sentence summary, adheres to a concise and direct style, and lacks redundant explanatory descriptions.
		\item Generated Intent Reasonableness: Whether the generated intent is reasonable and relevant.
	\end{itemize}
	
	For the recommendations following intent prediction, we also use five metrics for evaluation:
	
	\begin{itemize}
		\item Relevance: Whether the recommendation aligns with the user's intent.
		\item Usefulness: Whether the user would genuinely adopt the recommendation upon seeing it.
		\item Clarity: Whether the generated text is concise and easy to understand.
		\item Executability: Whether the suggestion can be clearly executed.
		\item Novelty/Surprise: Whether the suggestion offers new possibilities rather than merely repeating prior operations.
	\end{itemize}
	
	We employed a three-point scale (0–2) to score the aforementioned metrics:
	
	\begin{itemize}
		\item Score 2: Fully meets the evaluation criteria.
		\item Score 1: Partially meets the criteria.
		\item Score 0: Does not meet the core criteria.
	\end{itemize}

	Similar to many text generation tasks, due to the difficulty of evaluation and the limited reliability of automatic metrics, human evaluation is crucial. In our study, annotators review video frames alongside the ground truth task descriptions and complete two assessments:
	(1) Verify whether the trajectory in a given data sample is reasonable and practically useful, excluding noisy samples from the original dataset from evaluation.
	(2) Referring to the ground truth task description, score the various metrics for the video's predicted task description and suggestions.
	
	To evaluate the quality of our proposed task and metrics, we selected 30 high-quality data samples from the Agent-I dataset (containing 260 tasks automatically constructed by UI-Agent) and the Person-I dataset (consisting of 50 real-world scenarios) based on criteria (mainstream Apps, long length, fine-grained, multiple slots). This includes 25 single-App samples and 5 cross-App samples, used for prior intent summarization metrics. Additionally, two suggestions are obtained for each sample to measure operation and search suggestion metrics, ensuring reliable assessment of the suggestion metrics.
	
	Furthermore, we employ a Large Multimodal Model (LMM) as an automatic evaluator for assessing the task metrics. Recent advances in the LMM field have shown promising results in using LMMs to evaluate the task completion of autonomous agents \cite{he2024webvoyager,pan2024autonomous}. Based on this, we selected the latest model (Doubao-1.5-vision-pro-32k) as the automatic evaluator to assess the metric scores of the predicted task descriptions.
	
	\section{Experiments}
	\subsection{Experimental Settings}
	To validate the feasibility of MLLMs handling this task and the effectiveness of the two-stage filtering method, we integrated the task and algorithms into several widely-used, state-of-the-art multimodal baseline models, such as Doubao-1.5-vision-pro-32k (Main experimental model) , GPT-4o , and Qwen3-VL \cite{qwen3technicalreport}. In the experiments, we uniformly sampled frames from each video as input to the baseline models.
	For Qwen3-VL, we fine-tuned it to test whether its capabilities in intent summarization and prediction tasks could be improved on a smaller model, providing a reference for subsequent deployment on mobile devices.
	
	\subsection{Main Results}
	\subsubsection{Task Feasibility}
	We evaluated the performance of MLLM (Doubao-1.5-vision-pro-32k) in completing the three tasks, selecting 30 samples for prior intent summarization metrics and operation suggestion metrics, respectively. Four experts scored the results to ensure the reliability of the experiment, as shown in Table 1.
	
	\begin{table*}[h]
		\centering
		\caption{Prior Intent Summarization Metrics}
		\label{tab:example}
		\begin{tabularx}{\textwidth}{*{6}{>{\centering\arraybackslash}X}}
			\toprule
			& Action Info. Complete & Action Sequence Accurate & Object Detail Accurate & Output Format Standard & Generated Intent Reasonable \\
			\midrule
			Expert1 & 57 & 58 & 54 & 58 & 57 \\
			Expert2 & 56 & 55 & 50 & 60 & 55 \\
			Expert3 & 60 & 56 & 56 & 60 & 55 \\
			Expert4 & 58 & 58 & 58 & 58 & 49 \\
			Average & 0.96 & 0.96 & 0.91 & 0.98 & 0.90 \\
			LLM Evaluation & 41 & 55 & 47 & 57 & 57 \\
			\bottomrule
		\end{tabularx}
	\end{table*}

	Mainstream MLLMs achieved high scores on the intent summarization task, benefiting from their extensive training data and large parameter sizes. However, there is still room for improvement in the accuracy of object details and the reasonableness of the generated intent, which we will analyze in detail in the appendix. For the evaluation of intent summarization metrics, we introduced LMM for automated assessment. Experiments found that AI evaluation already possesses certain capabilities but still lags behind human evaluation, showing deviations in recognizing completeness, standardization, and detail accuracy. This result indicates the potential practical value of LMM-based automated evaluation in the development cycle, while also highlighting the necessity of human assessment.
	
	Regarding suggestion quality, we found that the scores for Usefulness and Novelty/Surprise were relatively low. The primary reason is that the model only accesses a limited amount of information, making it difficult to construct a complete user profile and, consequently, to generate personalized suggestions. This result indicates the future importance of building personalized user databases. Furthermore, the automation capabilities of UI-Agents will also influence improvements in this area.

	\begin{table*}[h]
		\centering
		\caption{Operation Suggestion Metrics}
		\label{tab:operation_suggestions}
		\begin{tabularx}{\textwidth}{*{6}{>{\centering\arraybackslash}X}}
			\toprule
			& Relevance & Usefulness & Clarity & Executability & Novelty/Surprise \\
			\midrule
			Expert1 & 119 & 96 & 117 & 117 & 85 \\
			Expert2 & 114 & 48 & 115 & 114 & 44 \\    
			Expert3 & 104 & 71 & 112 & 111 & 64 \\
			Expert4 & 97 & 70 & 105 & 97 & 63 \\
			Average & 0.90 & 0.59 & 0.94 & 0.91 & 0.53 \\
			\bottomrule
		\end{tabularx}
	\end{table*}
	
	\begin{table*}[h]
		\centering
		\caption{Search Suggestion Metrics}
		\label{tab:search_suggestions}
		\begin{tabularx}{\textwidth}{*{6}{>{\centering\arraybackslash}X}}
			\toprule
			& Relevance & Usefulness & Clarity & Executability & Novelty/Surprise \\
			\midrule
			Expert1 & 118 & 105 & 120 & 120 & 98 \\
			Expert2 & 115 & 82 & 112 & 110 & 73 \\    
			Expert3 & 103 & 94 & 116 & 112 & 91 \\
			Expert4 & 105 & 103 & 114 & 111 & 95 \\
			Average & 0.92 & 0.80 & 0.96 & 0.94 & 0.74 \\
			\bottomrule
		\end{tabularx}
	\end{table*}

	Additionally, we evaluated the intent summarization capabilities of other mainstream open-source models, including gemma-3n-E4B-it \cite{gemma_3n_2025}, MiniCPM-V 4.5 \cite{yao2024minicpm}, UI-TARS-1.5-7B \cite{qin2025ui}, and Kimi-VL-A3B-Thinking-2506 \cite{kimiteam2025kimivltechnicalreport}, using SBERT (sentence bert score) and ROUGE metrics for assessment.The SBERT \cite{reimers-2019-sentence-bert} is used to measure the sentence-level semantic similarity between the ground truth and the prediction, while ROUGE \cite{lin2004rouge} is used to measure the n-gram overlap between them.
	
	The results (Table 4) show that smaller parameter models possess a certain level of usability, and performance generally improves as model size increases, although they still lag behind larger parameter models. Since SBERT and ROUGE can reflect model performance to some extent, we will continue to use these two metrics as references in subsequent evaluations.

	\begin{table}[H]
		\centering
		\caption{Other open-source model performance.The SBERT score measures
			semantic similarity by embedding sentences into a vector space and
			computing cosine similarity. ROUGE scores evaluate summary quality through unigram overlap (ROUGE-1), bigram overlap (ROUGE-
			2), and the longest common subsequence (ROUGE-L), reflecting
			sentence structure.}
		\label{tab:input_modality}
		\begin{adjustbox}{max width=\linewidth}
			\begin{tabular}{l c c c c c}
				\toprule
				Model & Size (B) & SBERT & ROUGE-1 & ROUGE-2 & ROUGE-L \\
				\midrule
				gemma-3n& 4 & 0.61 & 0.34 & 0.16 & 0.26 \\
				MiniCPM& 8 & 0.69 & 0.44 & 0.23 & 0.35 \\
				UI-TARS & 7 & 0.70 & 0.39 & 0.24 & 0.36 \\
				Kimi-VL & 16-3 & 0.68 & 0.36 & 0.21 & 0.30 \\
				Doubao & - & 0.81 & 0.53 & 0.33 & 0.46 \\
				\bottomrule
			\end{tabular}
		\end{adjustbox}
	\end{table}

	\subsubsection{Correlation Analysis between Intent Summarization and Suggestions}
	Through linear regression analysis \cite{seber2003linear}, we found that the quality of intent summarization positively influences the utility of suggestions, which can be expressed by the formula: $Y = 0.4668X + 3.4225$, with a statistically significant p-value < 0.001.
	Therefore, evaluating the effectiveness of intent summarization can indirectly diagnose the effectiveness of intent suggestions. Since intent suggestions are difficult to evaluate automatically, subsequent comparative and ablation experiments will use the effectiveness of intent summarization to measure model capability.
	
	\subsubsection{Efficiency Analysis of Input Modalities}
	Given that the last frame often represents the user's final or primary intent, we aimed to investigate the effectiveness of using the full video, a sequence of keyframes, or just the last frame for intent prediction.
	We used SBERT and ROUGE to evaluate the alignment between the actual intent and the predicted intent summary, and employed expert human evaluation to score both the user trajectory summary and the predicted user intent summary. This analysis was conducted using 20 real user trajectories from the Person-I dataset.
	
	The experiments show that in the vast majority of cases, using a sequence of keyframes provides a significant improvement over using only the last frame. Furthermore, the keyframe sequence outperforms using the full video in terms of efficiency and some metrics, validating that removing and stitching redundant information does not reduce prediction accuracy. While the full video retains comprehensive information, leading to more complete user trajectory summaries, it may introduce redundant visual data that can cause a deviation in understanding the core intent for the intent summary task.
	
	We tested input widths of 512 and 384 pixels to observe the impact of different resolutions on summarization and analysis capabilities. A higher resolution (512px) aids in extracting and recognizing certain key information, reducing difficulties in identifying small fonts and UI components. The smaller 384px resolution, however, offers a 25\% speed increase. For the Qwen3-VL-2B model, we also tested a higher resolution of 886px, but the actual improvement was negligible and came with a significantly higher inference time cost (approximately 2.5 times that of 384px). Therefore, for subsequent fine-tuning, we used a moderate input width of 512 pixels.
	
	\begin{table}[H]
		\centering
		\caption{Input Modality Analysis (Qwen-2B-512px),the compression ratio here refers to the ratio of different input forms compared to the video (uniformly sampled frames)}
		\label{tab:input_modality}
		\begin{adjustbox}{max width=\linewidth}
			\begin{tabular}{l c c c c c}
				\toprule
				Input Modality & Compression (\%) & ROUGE-1 & ROUGE-2 & ROUGE-L & SBERT \\
				\midrule
				Full Video(Uniform sampling) & - & 0.53 & 0.36 & 0.47 & \textbf{0.77} \\
				Last Frame  & ~5 & 0.40 & 0.24 & 0.34 & 0.70 \\
				Keyframe Sequence & ~57 & \textbf{0.54} & \textbf{0.40} & \textbf{0.51} & 0.76 \\
				Keyframes with Stitching & ~44 & 0.50 & 0.35 & 0.46 & 0.76 \\
				\bottomrule
			\end{tabular}
		\end{adjustbox}
	\end{table}
	
	\begin{table}[H]
		\centering
		\caption{Input Modality Analysis (Qwen-2B-384px)}
		\label{tab:input_modality_384}
		\begin{adjustbox}{max width=\linewidth}
			\begin{tabular}{l c c c c c}
				\toprule
				Input Modality & Compression (\%) & ROUGE-1 & ROUGE-2 & ROUGE-L & SBERT \\
				\midrule
				Full Video(Uniform sampling) & - & 0.52 & 0.37 & 0.48 & 0.76 \\
				Last Frame  & ~5 & 0.38 & 0.20 & 0.32 & 0.66 \\
				Keyframe Sequence & ~57 & 0.47 & 0.33 & 0.44 & 0.75 \\
				Keyframes with Stitching & ~44 & 0.51 & 0.33 & 0.45 & 0.75 \\
				\bottomrule
			\end{tabular}
		\end{adjustbox}
	\end{table}
	
	The experiments reveal that the model can still produce relatively good results stably at a compression rate of around 50\%. Since the last frame typically captures the most important interface, the model can output relevant information with minimal input, but this is insufficient to support predicting the next action in most scenarios.
	
	We also compared the metric evaluations for suggestions generated based on the keyframe sequence versus the last frame only, as detailed in the table 7. Because the keyframe sequence can reconstruct the complete user trajectory and contains more information, it leads to significant improvements across metrics. This further justifies the rationality of using keyframe sequences over the last frame alone when computational resources permit. In this part of the experiment, we observed that operation suggestions, compared to search suggestions, are more specific and require more contextual information for support.
	
	\begin{table*}[h]
		\centering
		\caption{Suggestion Metrics: Keyframe Sequence vs Last Frame}
		\label{tab:suggestion_comparison}
		\begin{tabularx}{\textwidth}{l c c c c c}
			\toprule
			Suggestion Type  & Relevance & Usefulness & Novelty/Surprise & Clarity & Executability \\
			\midrule
			Operation Suggestions (Last Frame) & 18 & 16 & 5 & 40 & 40 \\
			Operation Suggestions (Keyframe Sequence) & 31 & 30 & 8 & 40 & 40 \\
			
			Improvement (\%) & 72.2 & 87.5 & 60.0 & - & - \\
			\midrule
			Search Suggestions (Last Frame) & 28 & 22 & 12 & 40 & 40 \\
			Search Suggestions (Keyframe Sequence) & 39 & 39 & 17 & 40 & 40 \\
			
			Improvement (\%) & 39.3 & 77.3 & 41.7 & - & - \\
			\bottomrule
		\end{tabularx}
	\end{table*}
	
	\subsubsection{Lightweight Model Performance Analysis}
	To meet the requirements for on-device deployment on mobile phones, we investigated the performance of a lightweight model (Qwen3-VL-2B) on this task. We conducted GRPO fine-tuning experiments \cite{deepseekai2025deepseekr1incentivizingreasoningcapability,hu2022lora} using the constructed AGENT-I dataset, utilizing 200 samples for the training set and 60 samples for the test set. The App scenarios in the training and test sets were non-overlapping to ensure generalization capability.
	
	By comparing the metrics with the base model, we observe that the fine-tuned model shows improvements across all performance metrics, especially in the ROUGE scores, and it can generate responses that better align with user needs.

	\begin{table}[H]
		\centering
		\caption{Qwen3-VL-2B-512-GRPO Fine-tuning Results}
		\label{tab:input_modality_384}
		\begin{adjustbox}{max width=\linewidth}
			\begin{tabular}{l c c c c c c }
				\toprule
				Model & SBERT & ROUGE-1 & ROUGE-2 & ROUGE-L & ROUGE Avg & Reward  \\
				\midrule
				BASE &0.825&	0.38&	0.22&	0.31&	0.30&	0.56  \\
				FT (Fine-Tuned)  &0.862&	0.48&	0.31&	0.40&	0.40&	0.66  \\
				Improvement (\%)& 4.48&26.32& 40.91& 29.03& 33.33& 17.86  \\
				\bottomrule
			\end{tabular}
		\end{adjustbox}
	\end{table}

	\subsection{Ablation Studies}
	This section validates the effectiveness of the model algorithm through multiple ablation experiments, all conducted under a training-free setting. The analysis focuses on two key dimensions: compression ratio and similarity measures.
	
	\subsubsection{Compression Ratio Analysis}
	Experiments investigate the impact of different compression ratios, revealing that the proposed method maintains robustness across a wide range of compression levels:
	When the compression ratio reaches 35-45\%, the model not only maintains performance but even surpasses the original baseline—showing improvement over uniform frame sampling on QwenVL.
	When the compression ratio increases to 50\%, performance remains largely comparable to the original model, indicating that the framework can withstand higher compression rates without significant performance degradation.
	However, when the compression ratio increases to 70\%, model performance declines somewhat but remains usable.
	
	Additionally, at the same compression ratio, incorporating stitching can restore a more complete trajectory, while keyframes provide more concise information. Overall, moderate compression achieves a good balance between efficiency and accuracy, whereas excessive compression inevitably harms performance. This conclusion offers practical guidance for setting compression ratios in real-world applications, which must balance computational efficiency with task performance.
	
	\begin{table}[H]
		\centering
		\caption{Compression Ratio Analysis}
		\label{tab:compression_analysis}
		\begin{adjustbox}{max width=\linewidth}
			\begin{tabular}{lccccc}
				\toprule
				Input Modality & Compression (\%) & ROUGE-1 & ROUGE-2 & ROUGE-L & SBERT \\
				\midrule
				Full video(Uniform sampling) & - & 0.52 & 0.37 & 0.48 & 0.76 \\
				Last Frame & 5 & 0.40 & 0.24 & 0.34 & 0.70 \\
				Keyframe Sequence & 30 & 0.48 & 0.31 & 0.41 & 0.74 \\
				Keyframe Sequence & 57 & \textbf{0.54} & \textbf{0.40} & \textbf{0.51} & 0.76 \\
				Keyframe Sequence & 64 & 0.53 & 0.34 & 0.48 & \textbf{0.77} \\
				Keyframes with Stitching & 30 & 0.46 & 0.30 & 0.41 & 0.75 \\
				Keyframes with Stitching & 44 & 0.50 & 0.35 & 0.46 & 0.76 \\
				Keyframes with Stitching & 49 & 0.52 & 0.36 & 0.48 & \textbf{0.77} \\
				\bottomrule
			\end{tabular}
		\end{adjustbox}
	\end{table}
	
	\subsubsection{Similarity Measure Analysis}
	To evaluate the impact of different similarity metrics in pixel compression, experiments compared L1 distance, phash+L1, CLIP(chinese-clip-vit-huge-patch14)\cite{chinese-clip}and phash+SSIM. The results are shown in the table below:  
	phash+SSIM achieved the best overall performance, while L1 distance performed slightly worse.  
	
	These results confirm that phash+SSIM is the most effective metric for guiding pixel pruning.  
	It aligns more closely with human perceptual decision-making: pHash provides a perceptually approximate global low-frequency consistency judgment, while SSIM further validates local structural consistency.  
	This two-step verification process better approximates human subjective judgments of whether two images are the same or similar.  
	In contrast, relying solely on L1 distance tends to bias tokens with larger magnitudes, and CLIP introduces nearly 1 billion parameters, which imposes a greater burden on edge devices without delivering superior performance.
	
	\begin{table}[H]
		\centering
		\caption{Similarity Measure Analysis}
		\label{tab:similarity_analysis}
		\begin{adjustbox}{max width=\linewidth}
			\begin{tabular}{lccccc}
				\toprule
				Similarity Measure & Compression (\%) & ROUGE-1 & ROUGE-2 & ROUGE-L & SBERT \\
				\midrule
				L1  & 54 & 0.51 & 0.35 & 0.46 & 0.76 \\
				L1  & 61 & 0.52 & 0.37 & 0.47 & \textbf{0.77} \\
				CLIP    &60&0.50&	0.33&	0.43&	0.73\\
				CLIP    &66&0.49&	0.35&	0.45&	0.74\\
				phash+L1 & 62 & 0.50 & 0.34 & 0.46 & 0.75 \\
				phash+L1 & 69 & 0.48 & 0.29 & 0.43 & 0.76 \\
				phash+SSIM & 57 & \textbf{0.54} & \textbf{0.40} & \textbf{0.51} & 0.76 \\
				phash+SSIM & 64 & 0.53 & 0.34 & 0.48 & \textbf{0.77} \\
				\bottomrule
			\end{tabular}
		\end{adjustbox}
	\end{table}

	\section{Conclusion and Future Work}
	This paper addresses the task of identifying user goals from UI interaction trajectories by designing a keyframe sequence algorithm that resolves issues such as video frame redundancy and excessive token consumption, thereby improving inference speed and accuracy. We conducted fine-grained evaluations for trajectory summarization and intent recognition, validating the current usability of MLLMs in these tasks and further extending the task boundaries by testing the feasibility of MLLMs in providing operational and search suggestions. The framework was deployed in real-world scenarios, enabling more diverse input methods for UI-Agents. Additionally, this work explores the potential performance improvements from fine-tuning techniques in trajectory summarization and intent recognition. Future research may include: constructing more personalized and long-term user memory banks, building sufficient datasets for training lightweight specialized models, and exploring token-level frame compression methods (though this may increase the computational load on mobile devices). These findings highlight the potential of our framework to advance the development of more efficient and concise UI understanding technologies and the integration of UI perception and agency.
	
	\section*{Limitations}
	Although our framework achieves efficient and concise integration of perception and agency, it still has the following limitations: 1. The dataset used in this study is primarily based on a Chinese environment; the suggestion and agency capabilities in English environments have not been tested. 2. Due to the lack of long-term real-user memory data for testing, some suggestions may struggle to achieve personalization and innovation. 3. The current work relies on the performance of existing large models and uses small-scale datasets for fine-tuning validation; achieving full scenario and task coverage would require larger-scale data collection efforts.
	
	\section*{Ethical Considerations}
	Despite the tremendous potential for innovation, the development of autonomous agents also entails important ethical considerations. Our research aims to infer user intent from UI interaction trajectories and generate corresponding suggestions; therefore, it is essential to acknowledge the potential privacy implications of tracking user behavior. Safeguarding the security and protection of related sensitive data is paramount. By implementing user opt-in consent, on-device processing, data anonymization, and other privacy-preserving techniques, we can mitigate risks while better protecting user data. Researchers and developers should proactively and forward-lookingly address these issues to build trust and promote responsible innovation in the field of autonomous agents.
	\normalsize
	\bibliography{main}
	
	\section*{Supplementary Material}
	\appendix
	
	\section{Datasets Overview}
	Due to the scarcity of prior research in this area, we constructed two new datasets (Agent-I and Person-I) for experimentation and testing.
	
	The Agent-I dataset is constructed from trajectories generated by an agent automatically completing instruction-based tasks. It can be viewed as the inverse process of GUI agent operation, and the intents in this dataset are relatively explicit. Given the current lack of UI trajectory datasets for mobile scenarios, particularly in Chinese, and the significant time required for manual collection, we followed the approach of \cite{berkovitch2025identifying} and utilized a UI Agent to build a collection of image-text pairs comprising UI trajectories obtained from user instructions (260 samples, covering 100+ App scenarios). For trajectory summarization and intent understanding tasks, since the latest large models already possess considerable analytical capability, we selected 30 high-quality tasks for scoring and testing. For the more challenging task of suggestion evaluation, we selected two suggestions per sample for scoring and testing to ensure reliability. This includes 25 single-App tasks covering: News, Shopping, Reading, Video, Healthcare, Finance, Food, Travel, Music, Office, Settings, and Concerts; and 5 cross-App tasks covering: Settings, Music, Travel, Study, and Shopping.
	
	The Person-I dataset is constructed by simulating user behavior in real-world scenarios to build task trajectories. Some of these trajectories have explicit intents, while others contain ambiguous intents. Since UI trajectories generated by a UI Agent may differ from the behavior of real users in authentic scenarios, we also collected 50 UI trajectories from real user scenarios for intent analysis and suggestion evaluation. The user intents were verified by experts to ensure accuracy.

	\section{Automated Evaluation and Human Evaluation}
	
	We used an LLM (Doubao-1.5-vision-pro-32k) to automatically evaluate the various metrics of the preceding intent summarization. The intent summarization performance of both Doubao-1.5-vision-pro-32k and GPT-4o was evaluated. As shown in the table 11 , Doubao achieved higher scores compared to GPT-4o, potentially due to its more recent training methods and better adaptation to the Chinese language context.
	
	\begin{table*}[h]
		\centering
		\caption{LLM Metrics for Preceding Intent Summarization}
		\label{tab:example}
		\begin{tabularx}{\textwidth}{*{6}{>{\centering\arraybackslash}X}}
			\toprule
			& Action Info. Complete & Action Sequence Accurate & Object Detail Accurate & Output Format Standard & Generated Intent Reasonable \\
			\midrule
			GPT-4o & 28 & 44 & 38 & 46 & 46 \\
			Doubao & 41 & 55 & 47 & 57 & 57 \\
			\bottomrule
		\end{tabularx}
	\end{table*}

	We tested the agreement between the Model and expert4 using Accuracy and Kappa metrics. The agreement rate between LLM and human ratings varied across different dimensions:
	Information completeness 46.7\% (k=0.223), order accuracy 90\% (k=0.710), detail completeness 66.7\% (k=0.325), format standardization 86.7\%, and intent reasonableness 66.7\% (k=0.275).
	The results indicate that this automated process achieved a moderate level of agreement with human evaluation, demonstrating a certain level of evaluation capability.
	In the two scenarios where the large model scored lower—detail completeness and intent reasonableness—we assessed the inter-rater agreement among different experts. Due to the inherent ambiguity of the task itself, there was also some disagreement among the experts, falling into the category of substantial agreement, as detailed in the table 12.

	\begin{table}[H]
		\centering
		\caption{Inter-Rater Agreement}
		\label{tab:lora_config_single}
		\setlength{\tabcolsep}{4pt}
		\begin{tabularx}{\columnwidth}{lXX}
			\toprule
			& Accuracy & Kappa \\
			\midrule
			\multicolumn{3}{l}{\textbf{Detail Completeness}} \\
			Expert1 vs Expert2 & 0.800 & 0.625 \\
			Expert1 vs Expert3 & 0.800 & 0.616 \\
			Expert1 vs Expert4 & 0.867 & 0.643 \\
			Expert2 vs Expert3 & 0.733 & 0.478 \\
			Expert2 vs Expert4 & 0.733 & 0.455 \\
			Expert3 vs Expert4 & 0.933 & 0.789 \\
			\midrule
			\multicolumn{3}{l}{\textbf{Intent Reasonableness}} \\
			Expert1 vs Expert2 & 0.800 & 0.333 \\
			Expert1 vs Expert3 & 0.867 & 0.556 \\
			Expert1 vs Expert4 & 0.667 & 0.275 \\
			Expert2 vs Expert3 & 0.867 & 0.676 \\
			Expert2 vs Expert4 & 0.733 & 0.510 \\
			Expert3 vs Expert4 & 0.600 & 0.265 \\
			\bottomrule
		\end{tabularx}
	\end{table}

	\section{Parameters and Reward Settings}
	We fine-tuned the model using LoRA with GRPO, with some key parameters shown in the table 13.
	
	\begin{table}[H]
		\centering
		\caption{LoRA Tuning Configuration}
		\label{tab:lora_config_single}
		\setlength{\tabcolsep}{4pt}
		\begin{tabularx}{\columnwidth}{lX}
			\toprule
			Hyper-parameter & Parameter Value \\
			\midrule
			\multicolumn{2}{l}{\textbf{LoRA}} \\
			LoRA Alpha & 32 \\
			LoRA Dropout & 0.05 \\
			LoRA Rank & 16 \\
			Target Modules & $q\_proj$, $k\_proj$, $v\_proj$, $o\_proj$, $gate\_proj$, $up\_proj$, $down\_proj$ \\
			\midrule
			\multicolumn{2}{l}{\textbf{Training}} \\
			Epochs & 2 \\
			Batch Size & 1 \\
			Learning Rate & $1\times10^{-5}$ \\
			Warmup Ratio & 0.03 \\
			\bottomrule
		\end{tabularx}
	\end{table}
	
	We primarily used GRPO to enhance the accuracy and conciseness of the outputs while standardizing the output format.
	The total reward consists of two parts:
	Semantic similarity reward similarity(pred, label) and
	Format compliance reward format(pred).
	The final total reward is:
	$w_{sim} * similarity + w_{fmt} * format$
	where $ w_{sim} = 0.8, w_{fmt} = 0.2$, and the result is clipped to the range [-1.0, 1.0].
	
	The semantic similarity is computed as a composite of SBERT and ROUGE scores. Semantic alignment is the primary focus, meaning the model can still achieve a high score even if it uses different phrasing to express the same intent.
	ROUGE serves as a safeguard; when semantic embeddings lead to misjudgment or are overly "vague," n-gram level overlap provides fine-grained textual proximity constraints, preventing the model from producing sentences that seem relevant but lack detail alignment.
	Format considerations include: 1. Length control (encouraging 20-50 characters, with mild bonuses for adjacent ranges, penalties for excessive length, to ensure concise outputs);
	2. Number of intra-sentence delimiters (to ensure conciseness);
	3. Presence of numbers (emphasizing the task's requirement to cover and be sensitive to numerical information);
	4. Presence of location keywords (emphasizing the task's requirement to cover and be sensitive to location information); along with corresponding rewards and penalties.
	
	Through GRPO, the model can effectively summarize intent trajectories and pay attention to certain key information. For example, for a task involving purchasing a medium cup of Boba milk tea, iced with 30\% sugar, on a food delivery platform, the original model failed to capture the user's personalized preference for "iced with 30\% sugar," whereas the fine-tuned model successfully noted that the user selected a medium cup, 30\% sugar, and iced.

	\section{Qualitative Analysis}
	\subsection{Analysis Summary of Multiple Model Outputs}
	This section analyzes the characteristics and issues present in the outputs of multiple models.
	
	The user performed the following operations: Opened the music ticketing App, viewed popular concerts in the Hangzhou area, clicked into the details page for the Leon 2025 Tour Concert on September 13, 2025, at the Hangzhou Olympic Sports Center Gymnasium, browsed details, popular reviews, notices, etc., viewed the ticket-purchasing guide, and then returned to the concert homepage.
	
	Doubao provided the analysis: \{"Operation":"Entered the concert page in the music ticketing app, viewed details and popular reviews for the 2025 Leon Tour Concert Hangzhou stop, clicked 'Book Now', read the ticket-purchasing guide","Intent":"Book tickets for the 2025 Leon Tour Concert Hangzhou stop"\}. It basically covers the operational trajectory and arrives at a relatively accurate intent summary (although the specific time is not given, this is acceptable since Leon only has one concert in Hangzhou, eliminating the need for time filtering).
	
	gemma-3n-E4B-it provided the analysis: \{"Operation": "The user browsed and selected performance information for the 2025 JVKE 'What Is Here' Return Concert Hangzhou stop on the music ticketing App, viewing details such as performance time, venue, ticket prices, popularity, and entered the purchase process.","Intent": "Purchase tickets for the 2025 JVKE 'What Is Here' Return Concert Hangzhou stop."\}. Because the music ticketing app homepage displayed a JVKE concert upon returning, Gemma's understanding deviated, focusing only on the final information without global consideration of the sequence.
	
	MiniCPM-V-4.5 (nothink) provided the analysis: \{"Operation":"In the music ticketing app, the user first selected the city as Hangzhou, after clicking the search button, viewed the concert category, selected the concert tag, browsed and clicked the details page for the 2025 Leon Tour Concert (Hangzhou Stop), viewed the performance introduction, user reviews, and ticket-purchasing guide, and finally prepared to purchase tickets for the Leon 2025 concert held at 19:00 on September 13th at the Hangzhou Olympic Sports Center Gymnasium.","Intent":"The user's main intent is to find and purchase tickets for the 2025 Leon Tour Concert Hangzhou stop, understand the performance details and user reviews, and complete the purchase."\}. It demonstrates good analytical capability but may produce redundant outputs or occasionally misidentify the core intent.
	
	UI-TARS-1.5-7B provided the analysis: \{"Operation": "Searched for and viewed detailed information about a Hangzhou concert in a ticketing app, including performance time, venue, ticket prices, performance introduction, and audience reviews, finally clicked the 'Grab Tickets Now' button, preparing to purchase tickets.", "Intent": "Purchase tickets for a Hangzhou concert"\}. Its analysis failed to capture crucial details such as the specific time and artist, which are particularly important for subsequent recommendations.
	
	\section{Error Analysis}
	This section provides further analysis of issues related to model intent understanding and suggestions.
	
	In terms of intent summarization and prediction, large-parameter models have already demonstrated satisfactory performance in most scenarios.However, these models tend to overlook details such as numerical filtering, leading to incomplete information recognition.For example, consider a user trajectory where the user searches for a computer brand on a shopping app and applies a price filter.The model output might be:{"operation":"Tap the shopping app on the phone's home screen, enter the brand name after accessing the homepage, and browse various types of laptops","intent":"Obtain information related to laptops"},which fails to mention the critical action of filtering and viewing the price range of 500–900 and the corresponding detailed parameters.
	\begin{figure*}[t] 
		\centering 
		\includegraphics[width=0.8\textwidth, keepaspectratio]{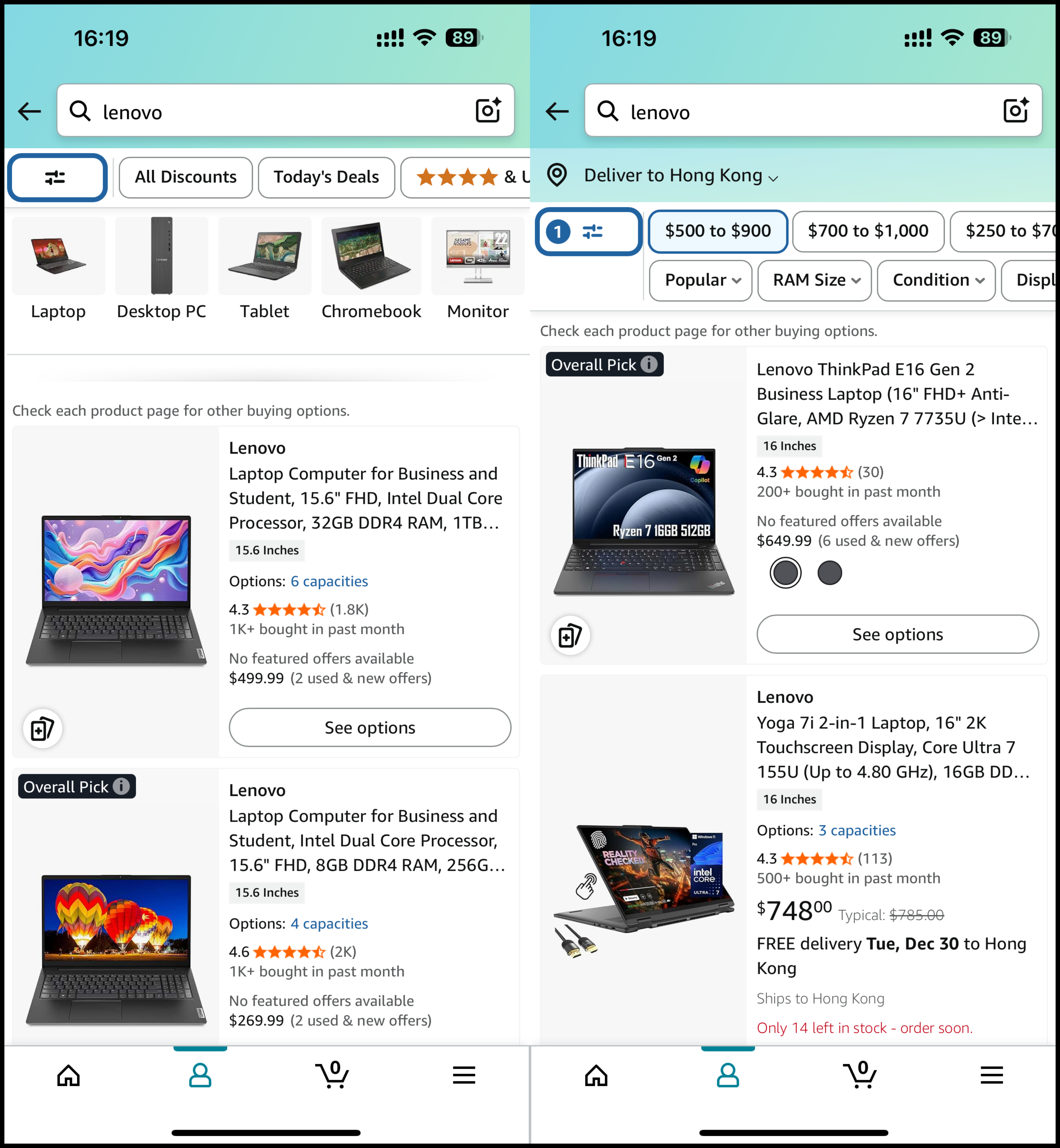}
		\caption{An example of a large model overlooking filtering information: a user checks laptops within the price range of 500–900, but the model output fails to mention this critical action and the detailed specifics of the target—the 500–900 price range.}
		\label{fig:flowchart}
	\end{figure*}
	
	Furthermore, both large and small parameter models share a common issue: they often pay attention to the initial gray placeholder text in search boxes, causing deviations in trajectory description and misidentification of the core intent. We categorize this as operational information recognition error. A summary of the encountered problems is presented in the table below.

	In terms of operational suggestions, models often struggle to recommend more in-depth filtering actions within the current context and tend to suggest cross-app operations instead.
	
	For example, in a “shopping scenario,” the model observes that the user searched for down jackets priced between 0–1000 RMB in a shopping app and filtered for Bosideng men’s items.
	The model then suggests search strategies across different platforms, such as opening another shopping app to search for “Bosideng men’s down jacket” and checking product details, or searching for “Bosideng men’s down jacket” in a secondhand marketplace app to compare used prices.
	While these suggestions are somewhat actionable, they are not sufficiently relevant or useful to the user’s current operating context. If the model relies only on the current frame, it also struggles to recognize the user’s emphasis on price, and may instead propose suggestions like searching “Bosideng down jacket” in the current app and reading user reviews—showing a mismatch with the user’s core intent.
	
	This difficulty in recommending deeper, more context-appropriate actions also appears in “information-seeking” scenarios. For instance, when a user is looking for pros and cons of law and accounting majors in a video app, the model suggests opening a college admissions app to check rankings for accounting, law, and English, or searching and opening related counseling posts in an image-and-text app to view major selection advice. However, it fails to offer more fitting suggestions within the current app, such as recommending high-quality video creators or related videos.
	
	We summarize the encountered problems in the table 14.
	
	\begin{table*}[h]
		\centering
		\caption{Summary of Issues in Intent Understanding and Next-Step Suggestions}
		\label{tab:combined_issues}
		\begin{tabularx}{\textwidth}{l>{\centering\arraybackslash}X>{\centering\arraybackslash}X}
			\toprule
			Category & Issue Type & Count \\
			\midrule
			\multirow{5}{*}{Intent Understanding} 
			& Missing Operational Info & 8 \\
			& Incorrect Operational Info & 2 \\
			& Key Info Not Specific & 3 \\
			& Incomplete/Vague Intent Summary & 8 \\
			& Image Compression Affects Summary & 1 \\
			\midrule
			\multirow{4}{*}{Next-Step Suggestions}
			& Suggestions Focus on Existing Features & 6 \\
			& Suggestions Lack Specificity & 2 \\
			& Search Not Based on Summarized Info & 2 \\
			& Actions/Search Ignore Personal Preference & 2 \\
			\bottomrule
		\end{tabularx}
	\end{table*}
	
	\section{Image Algorithm Examples}
	\subsection{Keyframe Selection}
	This section briefly demonstrates the effectiveness of frame selection. Between two similar frames, we tend to retain the latter frame because it often contains more information (such as additional user input). In Figure 3, the former frame has issues like UI widget not being fully loaded.
	\begin{figure*}[t] 
		\centering 
		\includegraphics[width=0.8\textwidth, keepaspectratio]{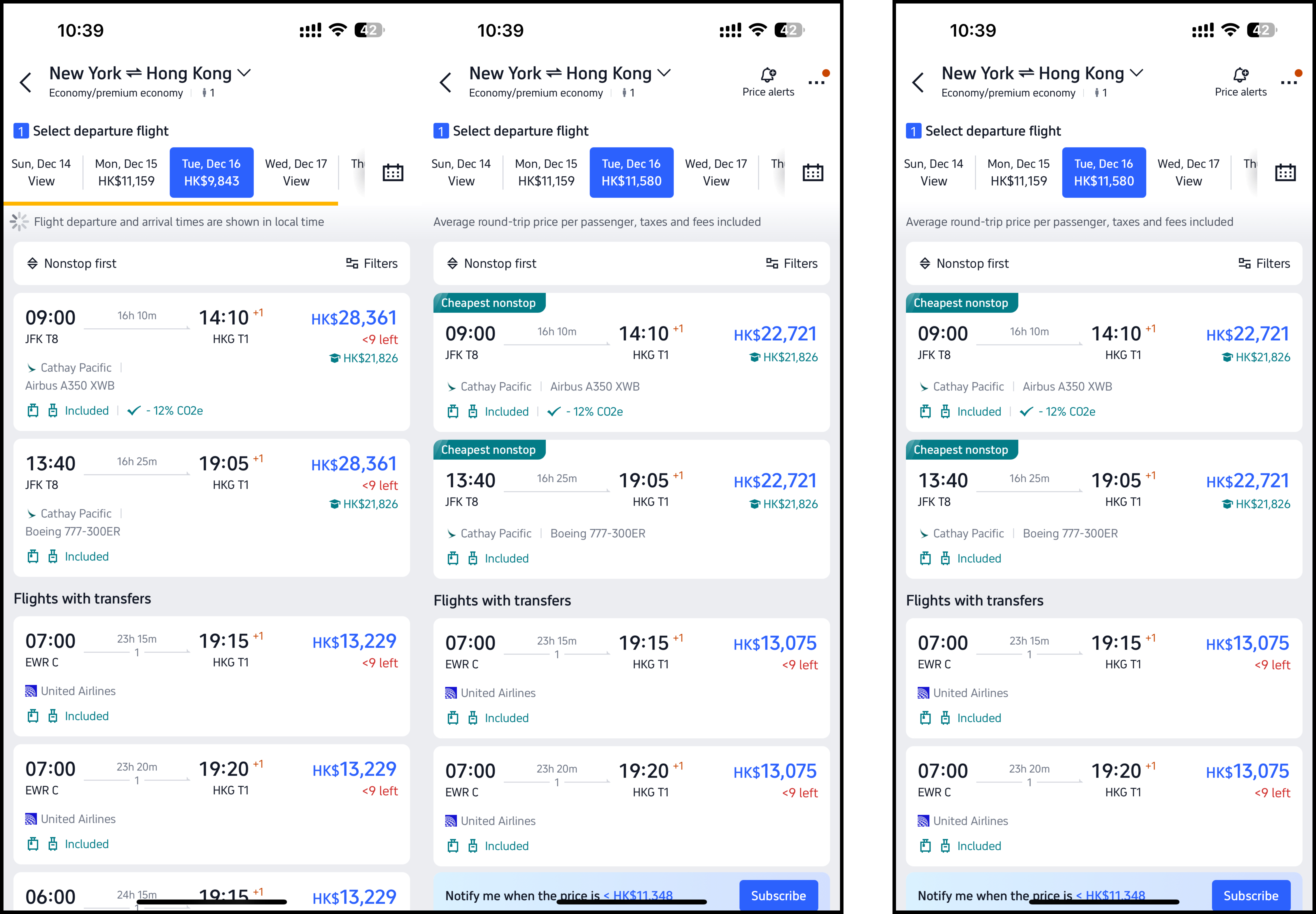}
		\caption{Keyframe Selection}
		\label{fig:keyframe_selection}
	\end{figure*}
	
	\subsection{Adaptive Stitching}
	This section briefly demonstrates the effectiveness of the stitching algorithm. Stitching can effectively reduce redundant pixel information caused by phone scrolling. As shown in the Figure 4, the stitching algorithm can reduce pixel redundancy by more than half in scrolling scenarios.
	\begin{figure*}[t] 
		\centering 
		\includegraphics[width=0.8\textwidth, keepaspectratio]{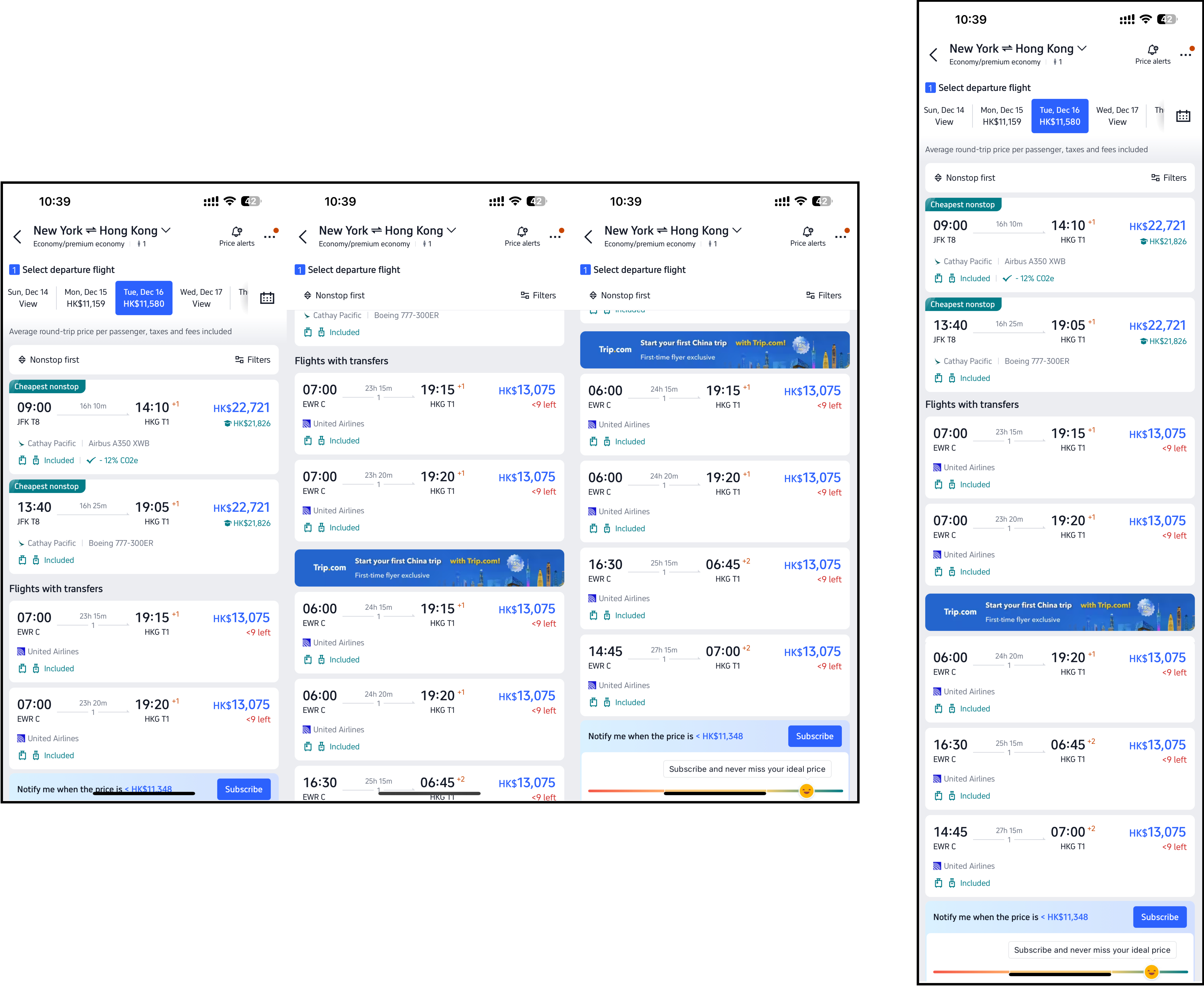}
		\caption{Adaptive Stitching}
		\label{fig:adaptive_stitching}
	\end{figure*}

	\section{Instructions and Prompts}
	This section presents the prompts we used, including trajectory intent summarization, search suggestions, operation suggestions, and LLM automated evaluation.
	
	\begin{figure*}[t] 
		\centering 
		\includegraphics[width=0.8\textwidth, keepaspectratio]{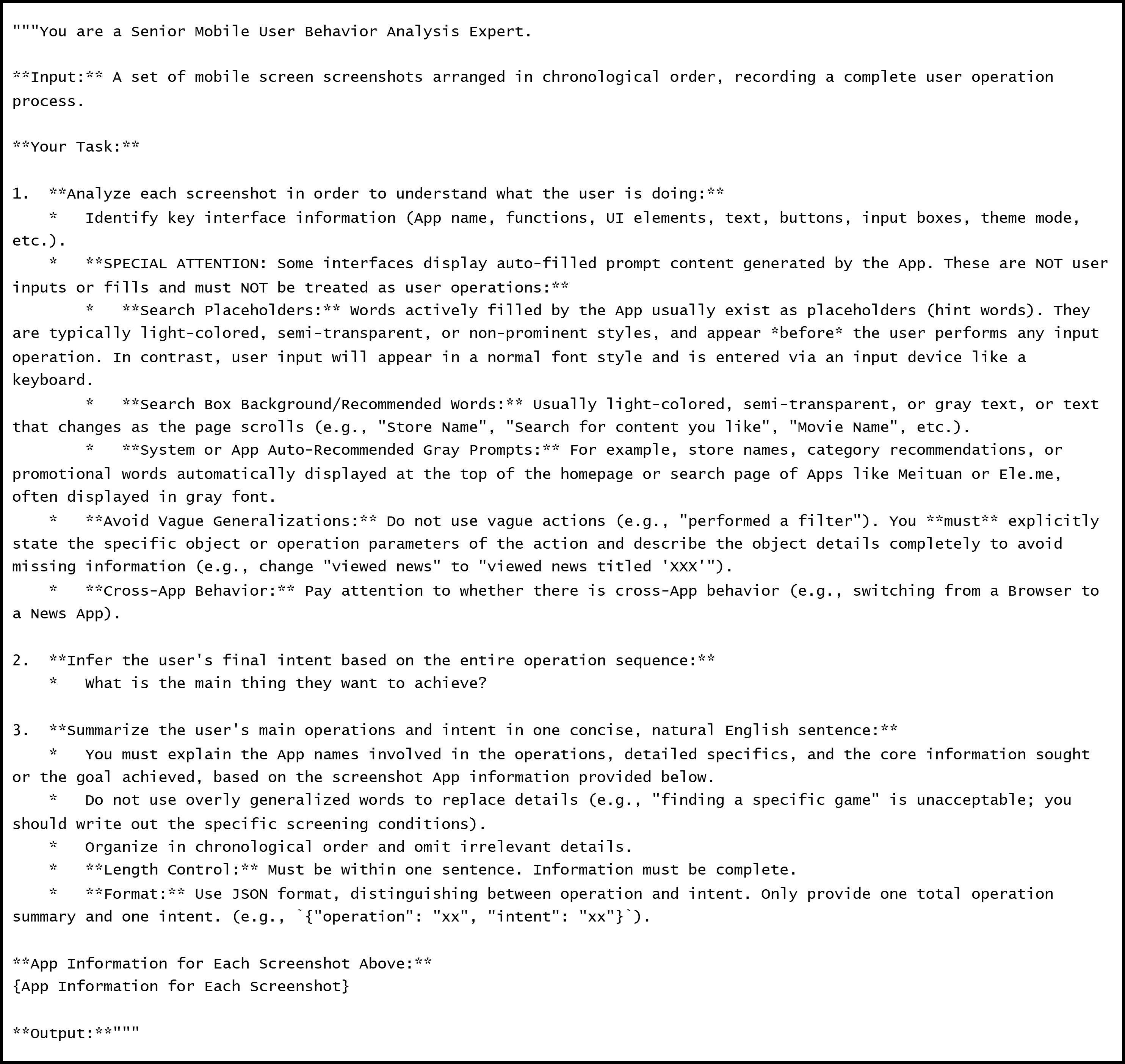}
		\caption{Prompt for intent summarization}
		\label{fig:flowchart}
	\end{figure*}
	
	\begin{figure*}[t] 
		\centering 
		\includegraphics[width=0.8\textwidth, keepaspectratio]{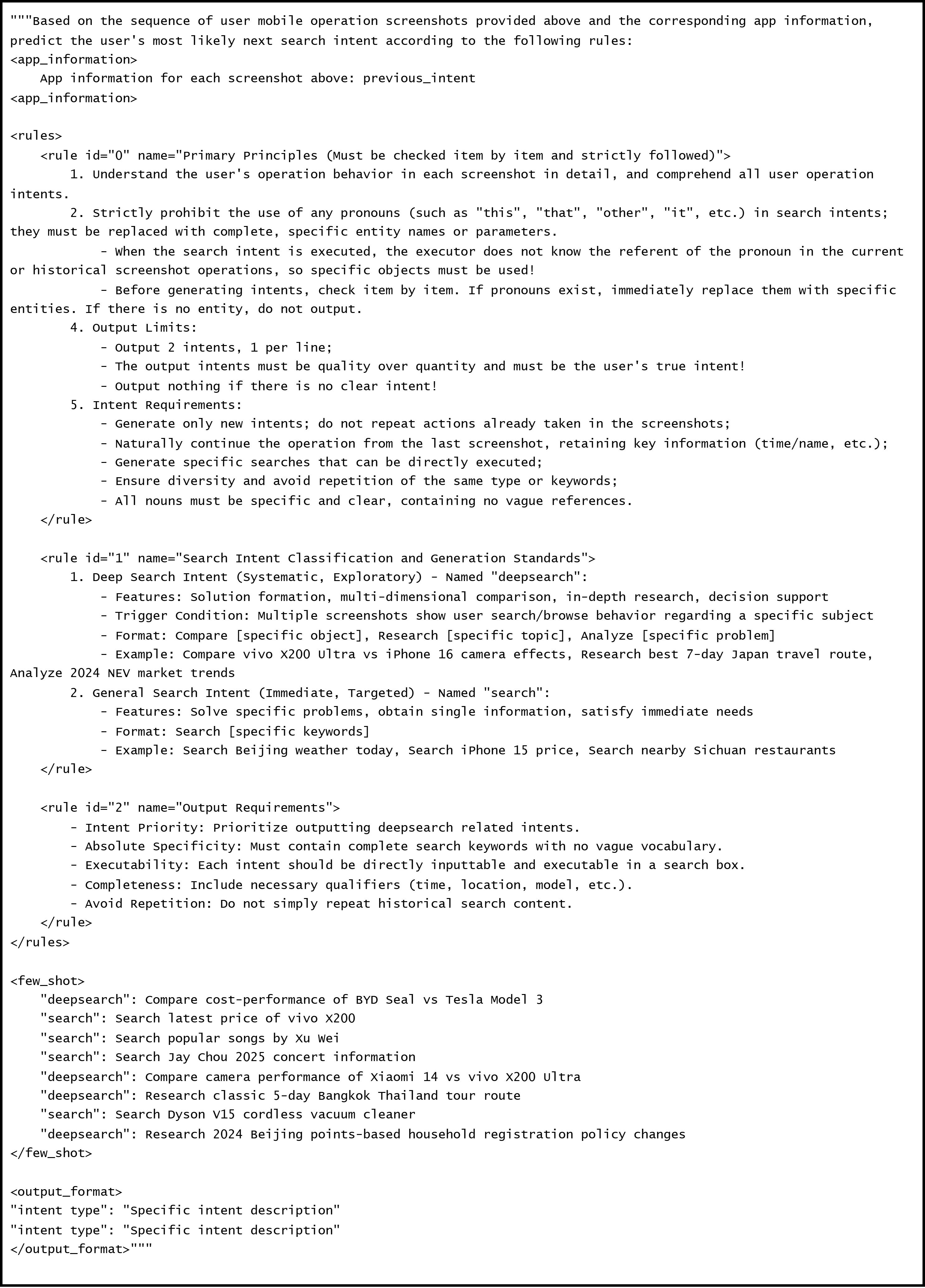}
		\caption{Prompt for search suggestions}
		\label{fig:flowchart}
	\end{figure*}	
	
	\begin{figure*}[t] 
		\centering 
		\includegraphics[width=0.8\textwidth, keepaspectratio]{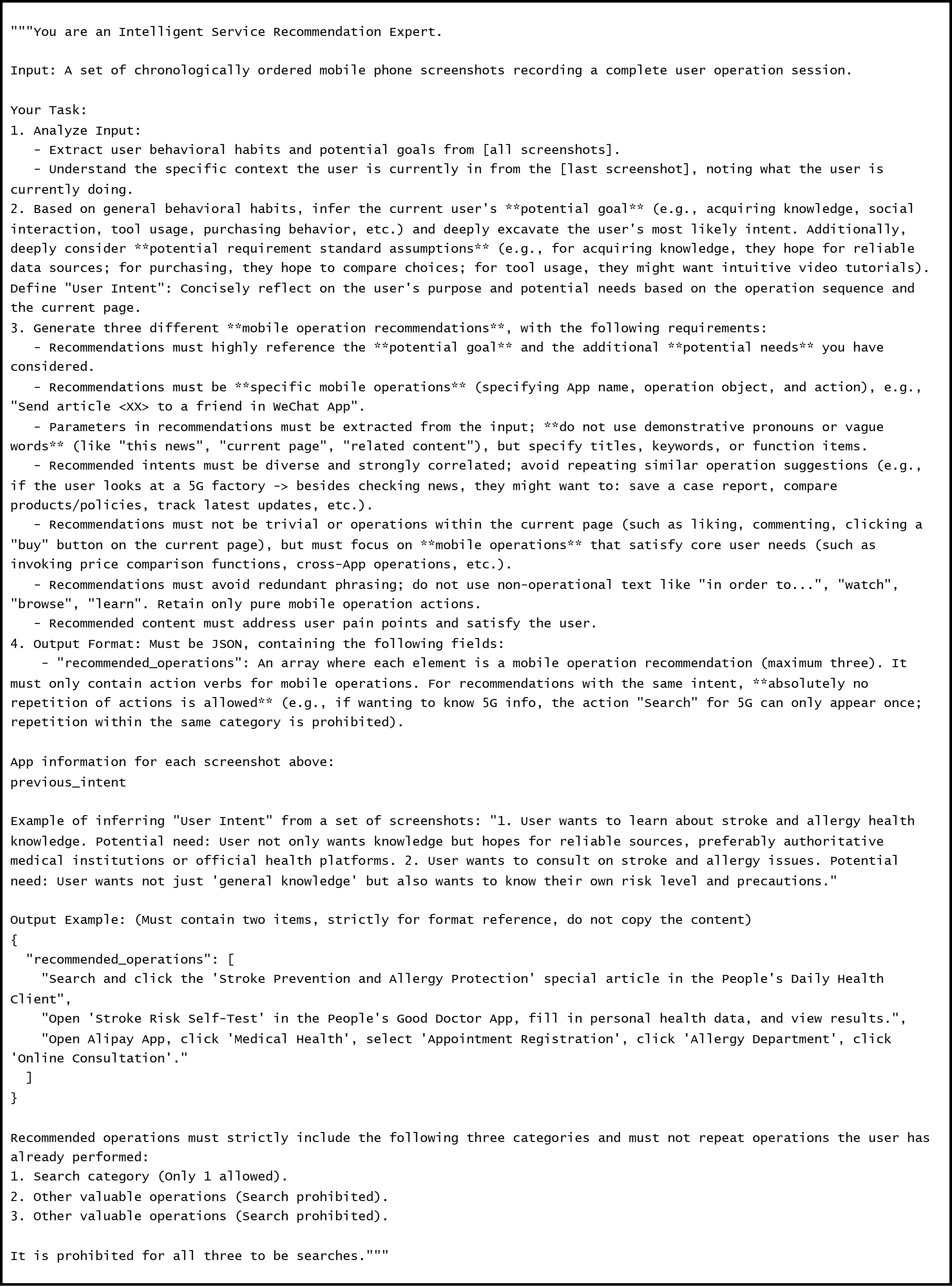}
		\caption{Prompt for operate suggestions}
		\label{fig:flowchart}
	\end{figure*}	
	
	\begin{figure*}[t] 
		\centering 
		\includegraphics[width=0.8\textwidth, keepaspectratio]{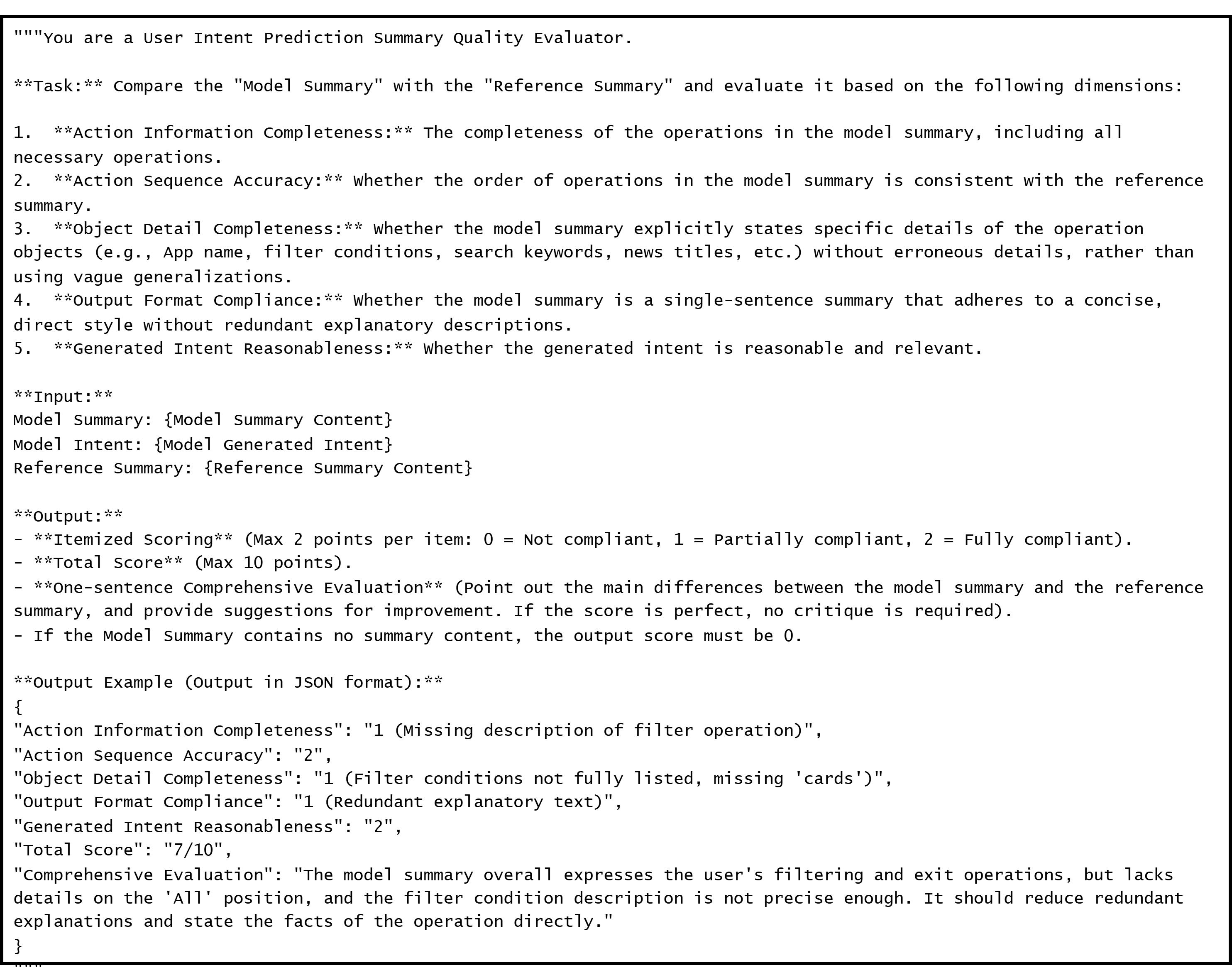}
		\caption{Prompt for automated evaluation of summary quality}
		\label{fig:flowchart}
	\end{figure*}

	\begin{figure*}[t] 
		\centering 
		\includegraphics[width=0.8\textwidth, keepaspectratio]{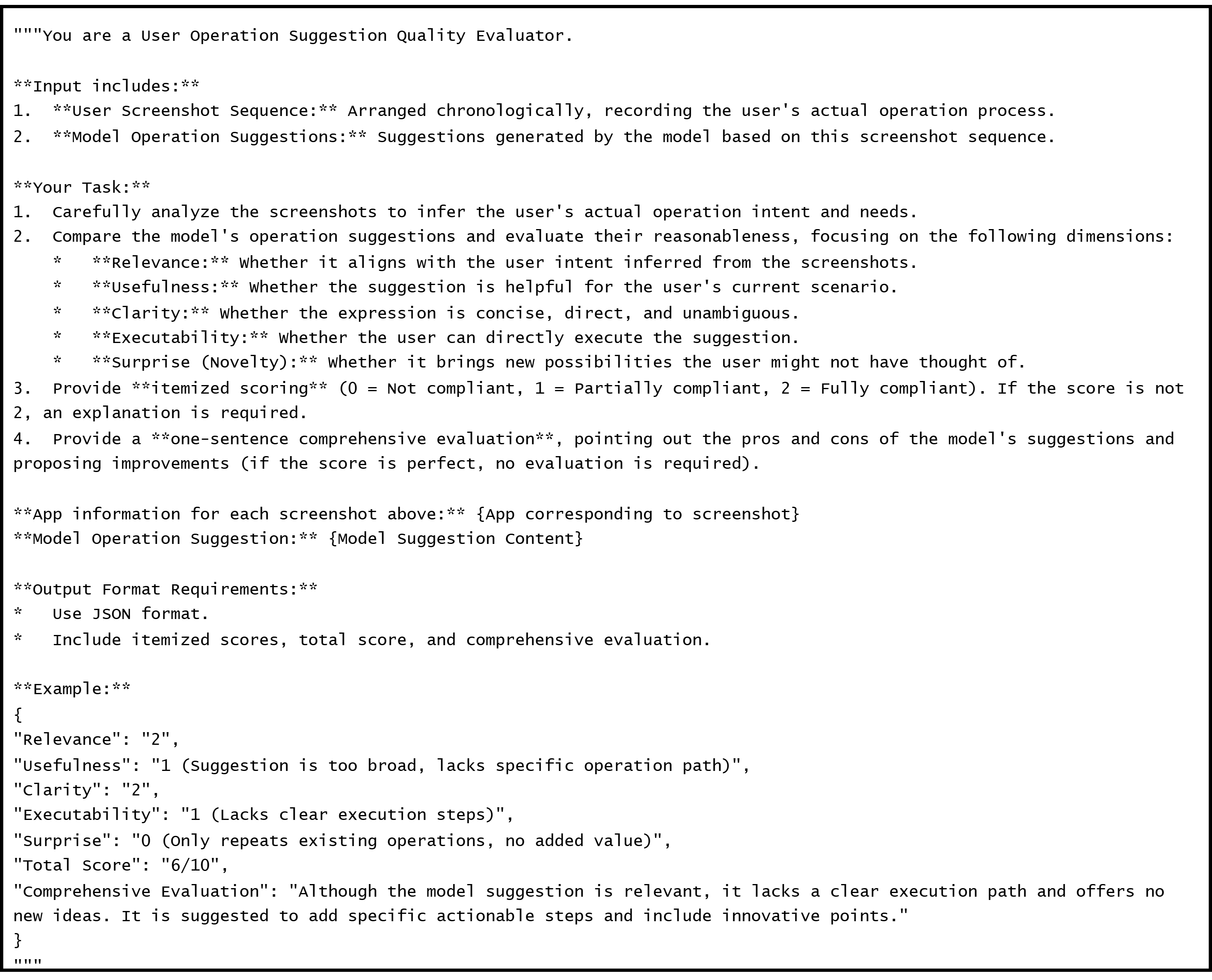}
		\caption{Prompt for automated evaluation of operation-suggestions quality}
		\label{fig:flowchart}
	\end{figure*}

\end{document}